\documentclass[10pt,twocolumn]{article}

\usepackage[capitalise]{cleveref} 
\usepackage{iccv}
\usepackage{times}
\usepackage{epsfig}
\usepackage{graphicx}
\usepackage{amsmath}
\usepackage{amssymb}
\usepackage{subcaption}
\usepackage{rotating}
\usepackage{multirow}
\usepackage{hhline}
\usepackage{float}
\usepackage{placeins}

\usepackage[binary-units,group-separator={,}, detect-weight=true]{siunitx}
\DeclareSIUnit\pixel{px}

\newcommand{\res}[1]{{$\SI{#1}{\percent}$}}
\newcommand{\bres}[1]{\textbf{{\SI{#1}{\percent}}}}

\newcommand{\kitti}{KITTI}

\crefname{table}{Table}{Tables}
\crefname{figure}{Figure}{Figures}

\usepackage[dvipsnames]{xcolor}

\usepackage{booktabs}

\usepackage[pagebackref=true,breaklinks=true,colorlinks,bookmarks=false]{hyperref}

\iccvfinalcopy 

\def\iccvPaperID{****} 

\begin{document}

\title{MultiNet: Real-time Joint Semantic Reasoning for Autonomous Driving}

\author{Marvin Teichmann\textsuperscript{1}\textsuperscript{2}\textsuperscript{3}, Michael Weber\textsuperscript{2}, Marius Z\"{o}llner\textsuperscript{2}, Roberto Cipolla\textsuperscript{3} and Raquel Urtasun\textsuperscript{1}\textsuperscript{4}\\[1em]
\textsuperscript{1} Department of Computer Science, University of Toronto \\
\textsuperscript{2} FZI Research Center for Information Technology, Karlsruhe\\
\textsuperscript{3} Department of Engineering, University of Cambridge\\
\textsuperscript{4} Uber Advanced Technologies Group\\[0.5em]
{\tt\small marvin.teichmann@googlemail.com, Michael.Weber@fzi.de,} \\[-0.3em]
{\tt\small zoellner@fzi.de, rc10001@cam.ac.uk, urtasun@cs.toronto.edu}
}

\maketitle

\begin{abstract}

While most approaches to semantic reasoning have  focused on improving performance, in this paper we argue that computational times are very important in order to enable real time applications such as autonomous driving. 
Towards this goal, we present an approach to joint classification, detection and semantic segmentation via a  unified architecture where the encoder is shared amongst the three tasks.
Our approach is very simple, can be trained end-to-end and performs extremely well in the challenging KITTI dataset, outperforming the state-of-the-art in  the road segmentation task. Our approach is also very efficient, allowing us to perform inference at  more then 23 frames per second.

Training scripts and trained weights to reproduce our results can be found here: \url{https://github.com/MarvinTeichmann/MultiNet}


%
%
%

\end{abstract}
\section{Introduction}
\label{sec:Introduction}

Current advances in the field of computer vision have made clear that visual perception is going to play a key role in the development of self-driving cars. This is mostly due to the deep learning revolution which begun with the introduction of AlexNet in 2012 \cite{NIPS2012_4824}. Since then, the accuracy of new approaches has been increasing at a vertiginous rate. Causes of this are the existence of more data, increased computation power and algorithmic developments.  The current trend is to create deeper networks with as many layers as possible \cite{DBLP:journals/corr/HeZRS15}. 
 
While performance is already extremely high, when dealing with real-world applications, running times becomes important. 
New hardware accelerators as well as compression, reduced precision and distillation methods have been exploited to speed up current networks. 

\begin{figure}[t]
    \centering
    \includegraphics[width=\columnwidth]{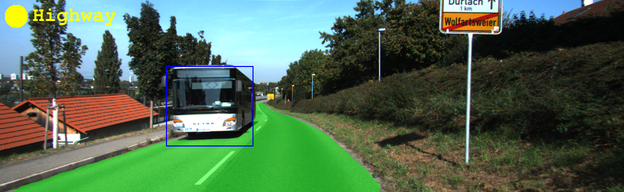}
    \caption{Our goal: Solving street classification, vehicle detection and road segmentation in one forward pass.}
    \label{fig:intro} 
\end{figure} 

In this paper we take an alternative approach and design a network architecture that can very efficiently perform  classification, detection and semantic segmentation simultaneously. 
This is done by incorporating all three task into a unified encoder-decoder architecture. We name our approach MultiNet. 

The encoder is a deep CNN, producing rich features that are shared among all task. Those features are then utilized by task-specific decoders, which produce their outputs in real-time.
In particular, the detection decoder combines the  fast regression  design introduced in Yolo \cite{DBLP:journals/corr/RedmonDGF15} with the size-adjusting ROI-align  of Faster-RCNN \cite{DBLP:journals/corr/Girshick15}and Mask-RCNN \cite{DBLP:journals/corr/HeGDG17}, achieving a better speed-accuracy ratio.

We demonstrate the effectiveness of our approach in the challenging KITTI  benchmark \cite{Geiger2012CVPR} and show state-of-the-art performance in road segmentation. Importantly, our  ROI-align implementation can significantly improve detection performance without requiring an explicit proposal generation network. This gives our decoder a significant speed advantage compared to Faster-RCNN \cite{DBLP:journals/corr/RenHG015}. Our approach is able to benefit from sharing computations, allowing us to perform inference in less than 45 ms for all tasks.

All our code, training scripts and weights, required to reproduce our results, are released on Github.



\begin{figure*}
\centering
  \includegraphics[width=\textwidth]{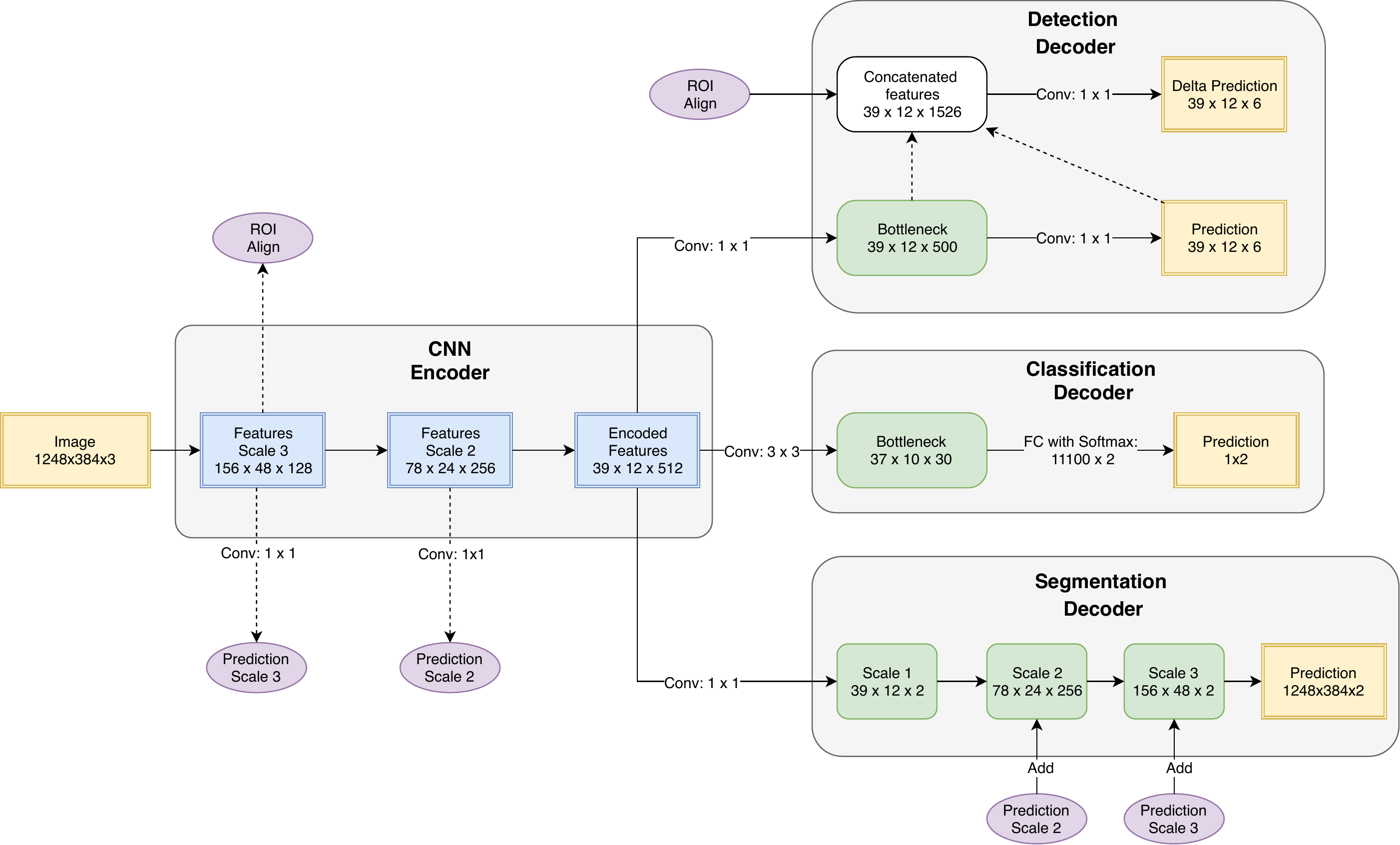}
  \caption{MultiNet architecture.}
  \label{fig:MultiNet}
\end{figure*}

\section{Related Work}
\label{sec:related}

In this section we review  current approaches to the tasks that MultiNet tackles, i.e., detection, classification and semantic segmentation. We focus our attention on deep learning based approaches. 


\paragraph{Classification:} After the development of AlexNet \cite{NIPS2012_4824}, most  modern approaches to image classification utilize deep learning. 
Residual networks \cite{DBLP:journals/corr/HeZRS15} constitute the state-of-the-art, as they allow to train very deep networks without problems of vanishing or exploding  gradients. 
In the context of road classification, deep neural networks are also widely employed \cite{ma2016find}. Sensor fusion has also been exploited in this context \cite{seegertowards}. In this paper we use classification to guide other semantic tasks, i.e., segmentation and detection. 

\paragraph{Detection:} Traditional deep learning approaches to object detection  follow a two step process, where  region proposals \cite{lampert2008beyond,DBLP:journals/corr/HosangBS14,DBLP:journals/corr/HosangBDS15} are first generated and  then scored using a convolutional network   \cite{DBLP:journals/corr/GirshickDDM13,DBLP:journals/corr/RenHG015}.  
Additional performance improvements can be gained by  using convolutional neural networks (CNNs) for the proposal generation step \cite{DBLP:journals/corr/ErhanSTA13,DBLP:journals/corr/RenHG015} or by reasoning in 3D \cite{chen20153d,chen2016monocular}. 
Recently, several  methods have  proposed to use a single deep network that is trainable end-to-end to directly perform detection \cite{DBLP:journals/corr/SermanetEZMFL13, DBLP:journals/corr/LiuAESR15,stewart2016end, DBLP:journals/corr/LiuAESR15}. Their main advantage over proposal-based methods is that they are much faster at both training and inference time, and thus more suitable for real-time detection applications. However, so far they lag far behind in performance. 
In this paper we propose an end-to-end trainable detector which reduces significantly the performance gap. We argue that the main advantage of proposal-based methods is their ability to have size-adjustable features. This inspired our ROI pooling implementation.



\paragraph{Segmentation:} 
Inspired by the successes of deep learning,  CNN-based classifiers were adapted to the task of semantic segmentation. Early approaches used the inherent efficiency of CNNs to implement implicit sliding-window \cite{fast_scanning,highly}. 
 FCN were proposed to model semantic segmentation using a deep learning pipeline that is trainable end-to-end. Transposed convolutions \cite{zeiler2010deconvolutional,dumoulin2016guide,DBLP:journals/corr/ImKJM16} are utilized to upsample low resolution features. A variety of deeper flavors of FCNs have been proposed since \cite{DBLP:journals/corr/BadrinarayananK15,noh2015learning,DBLP:journals/corr/RonnebergerFB15,googleSeg}. Very good results are archived by combining FCN with conditional random fields (CRFs) \cite{CRF1,CRF2,DBLP:journals/corr/ChenPK0Y16}.  \cite{CRF1,DBLP:journals/corr/SchwingU15} showed that mean-field inference in the CRF can be cast as a recurrent net allowing end-to-end training.
Dilated convolutions were introduced in  \cite{DBLP:journals/corr/YuK15} to augment the receptive field size without losing resolution. 
The aforementioned techniques in conjunction with residual networks \cite{DBLP:journals/corr/HeZRS15} are currently the state-of-the-art. 

\paragraph{Multi-Task Learning:} 

Multi-task learning techniques aim at learning better representations by exploiting many tasks. Several approaches have been proposed in the context of  CNNs \cite{DBLP:journals/corr/Long015a,liu2015representation}. An important application for multi-task learning is face recognition \cite{zhang2014facial,yim2015rotating,DBLP:journals/corr/RanjanPC16}.

Learning semantic segmentation in order to perform detection or instance segmentation has been studied \cite{gidaris2015object, dai2016instance, pinheiro2016learning}. In those systems, the main goal is to perform an instance level task. Semantic annotation is only viewed as an intermediate result. Systems like \cite{DBLP:journals/corr/SermanetEZMFL13,DBLP:journals/corr/WuSH16e} and many more design one system which can be fine-tuned to perform tasks like classification, detection or semantic segmentation. In this kind of approaches, a different set of parameters is learned for each task. Thus, joint inference is not possible in this models. The system described in \cite{hariharan2014simultaneous} is closest to our model. However this system relies on existing object detectors and does not fully leverage the rich features learned during segmentation for both tasks. To the best of our knowledge our system is the first one proposed which is able to do this.

\section{MultiNet for Joint Semantic Reasoning }
\label{sec:architecture}

In this paper we propose an efficient and effective feed-forward architecture, which we call {\it MultiNet}, to jointly reason about semantic segmentation, image classification and object detection. 
Our approach shares a common encoder over the three tasks and has three branches, each implementing  a decoder for a given task. 
We refer the reader to Fig. \ref{fig:MultiNet} for an illustration of our architecture.
MultiNet can be trained end-to-end and joint inference over all tasks can be done in less than 45ms. We start our discussion by introducing our joint encoder, followed by the task-specific decoders.


\subsection{Encoder} 
\label{sec:encoder}

The task of the encoder is to process the image and  extract rich abstract features \cite{zeiler2014visualizing} that contain all necessary information to perform accurate segmentation, detection and image classification. 
The encoder consists of the convolutional and pooling layers of a classification network. The weights of the encoder are initialized using  the weights pre-trained on ImageNet Classification Data \cite{ILSVRC15}. As encoder any modern classification network can be utilized.

We perform experiments using versions of VGG16 \cite{zeiler2014visualizing} and ResNet \cite{DBLP:journals/corr/HeZRS15} architectures. Our first VGG encoder uses all convolutional and pooling layers of VGG16. but discards the fully-connected softmax layers. We call this version \emph{VGG-pool5}, as pool5 is the last layer used from VGG16. The second implementation only discards the final fully-connected softmax layer. We call this architecture \emph{VGG-fc7}, as fc7 is the last layer used from VGG16. VGG-fc7 utilizes two fully-connected layers from VGG, namely \emph{fc6} and \emph{fc7}. We replace those layers with equal $1 \times 1$ convolutions as discussed in \cite{DBLP:journals/corr/SermanetEZMFL13,long2015fully}. This trick allows the encoder to process images with arbitrary input size. In particular we are not bound to the original VGG input of $224 \times 224$, which would be to small to perform perception in street scenes.

For ResNet we implement the $50$ and $101$ layer Version of the Network. As encoder we utilize all layers apart from the layers fully-connected softmax.

\begin{figure}
    \centering
    \includegraphics[width=\columnwidth]{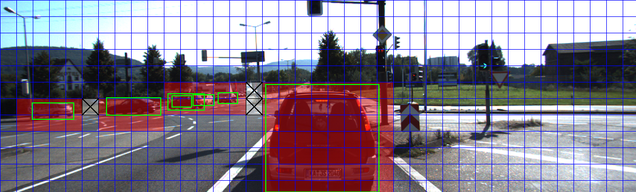}
    \caption{Visualization of our detection encoding. Blue grid: cells, Red cells: cells with positive confidence label. Transparent Cells: cells with negative confidence label. Grey cells: cells in don't care area. Green boxes: ground truth boxes.}
    \label{fig:cells} 
\end{figure}



\subsection{Classification Decoder}
\label{sec:class_decoder}


We implement two classification decoders. One version is a vanilla fully-connected layer with softmax activation. This encoder is used in conjunction with an input size of $224 \times 224$. Thus, the overall network is equal to the original VGG or ResNet respectively, when used with the corresponding encoder. The purpose of this encoder is to serve as high quality baseline to show the effectiveness of our scene classification approach. This first classification encoder cannot be used for joint inference with segmentation and detection. Both approaches require a larger input size. Increasing the input size on this classification encoder however, yields into an unreasonable high amount of parameters for the final layer.

The second classification decoder is designed to take advantage of the high resolution features our encoder generates. In typical image classification tasks (e.g. \cite{ILSVRC15,cifar}) the input features one object, usually centred prominently in the image. For this kind of task it is reasonable to use a very small input size. Street scenes on the other hand contain a large amount of smaller scale objects. We argue that it is vital to use high-resolution input in order to utilize features those objects provide. By increasing the input size of our image to $1248 \times 348$, we effectively apply our feature generator to each spatial location of the image \cite{DBLP:journals/corr/SermanetEZMFL13,long2015fully}. The result is a grid of $39 \times 12$ features, each corresponding to a spatial region of size $32 \times 32$ pixels. In order to utilize this features, we first apply a $1 \times 1$ convolution with $30$ channels. This layer serves as \emph{BottleNeck}. The main purpose is to greatly reduce dimensionality.

\subsection{Detection Decoder}

The detection decoder is designed to be a proposal free approach similar to ReInspect \cite{stewart2016end}, Yolo \cite{DBLP:journals/corr/RedmonDGF15} and Overfeat \cite{DBLP:journals/corr/SermanetEZMFL13}. By omitting and artificial proposal generator step much faster inference can be obtained. This is crucial towards our goal of building a real-time capable detection system. 

Proposal based detection systems have a crucial advantage over non-proposal based. They internally rescale the rich features utilized for detection. This makes the CNN internally invariant to scale. This is a crucial feature, as CNN are naturally not able to generalize over different scales. We argue, that the scale invariance is the main advantage of proposal based systems. 

Our detection decoder tries to close the marry the good detection performance of proposal based detection systems with the fast speed of non-proposal based systems. To archive this, we include a rescaling layer inside the decoder. The rescaling layer consists of RoI align \cite{DBLP:journals/corr/HeGDG17} and provides the main advantage of proposal based systems. Unlike proposal based systems, no non-differential operations are done and the rescaling can be computed very efficiently.

The first step of our decoder is to produce a rough estimate of the bounding boxes. Towards this goal, we first pass the encoded features through a $1\times 1$ convolutional layer with 500 filters, producing a tensor of shape $39\times12\times500$. Those features serve as \emph{bottleneck}. This tensor is processed with another $1\times 1$  convolutional layer which outputs 6 channels at resolution $39\times12$. We call this tensor \emph{prediction}, the values of the tensor have a semantic meaning. The first two channels of this tensor form a coarse segmentation of the image. Their values represent the confidence that  an object of interest is present at that particular location in the $39\times12$ grid. The last four channels represent the coordinates of a bounding box  in the area around that cell. Fig. \ref{fig:cells} shows an  image with its cells.

Those prediction are then utilized to introduce scale invariance. A rescaling approach, similar to the ones found in proposal based systems is applied on the initial coarse prediction. The rescaling layer follows the RoI align strategy of \cite{DBLP:journals/corr/HeGDG17}. It uses however the prediction of each cell to produce a RoI align. This makes the operation differentiable. Thus it can be implemented insight the CNN pooling. The result is an end-to-end trainable system which is faster. The features pooled by the RoI align are concatenated with the initial prediction and used to produce a more accurate prediction. The second prediction is modeled as offset, its output is added to the initial prediction.

\subsection{Segmentation Decoder}

The segmentation decoder follows the main ideas of the FCN architecture \cite{long2015fully}. Given the features produced by the encoder, we produce a low resolution segmentation of size $39 \times 12$ using a $1\times 1$ convolutional layer. This output is then upsampled using three transposed convolution layers \cite{dumoulin2016guide}. Skip connections are utilized to extract high resolution features from the lower layers. Those features are first processed by a $1 \times 1$ convolution layer and then added to the partially upsampled results.

%

%

\section{Training Details}
\label{sec:training}

In this section we  describe the loss functions we employ as well as other details of our training procedure including initialization.

\paragraph{MultiNet Training Strategy:}

MultiNet training follows a fine-tuning approach. First the encoder network is trained to perform classification on the ILSVRC2012 \cite{5206848} data. In practice, this step is omitted. Instead we initialize the weights of all layers of the encoder with weights published by the authors whose network architecture we are using.

In a second step, the final fully connected layers are removed and replaced by our decoders. Then the network is trained end-to-end using KITTI data. Thus MultiNet training follows a classic fine-tuning pipeline. 

Our joint training implementation computes the forward passes for examples corresponding to each of the three tasks independently. The gradients are only added during the back-propagation steps. This has the practical advantage that we are able to use different training parameters for each decoder. Having this degree of freedom is an important feature of our joint training implementation. The classification task for example requires a relative large batch size and more aggressive data-augmentation than the segmentation task to perform well.

\paragraph{Loss function:}  Classification and segmentation are trained using a softmax cross-entropy loss function.

For the detection, the final prediction is a grid of $12 \times 39$ cells. Each cell gets assigned a confidence label as well as a box label. The box label encodes the coordinates of the box and is parametrized relative to the position of a cell. A cell $c$ gets assigned a positive confidence label if and only if it intersects with at least one bounding box. If this is the case the cell also gets assigned to predict the coordinates of the box it intersects with. If multiple boxes intersect with a cell, the box whose centre is closest to the centre of $c$ is chosen. Note that one box can be predicted by multiple cells.

If a box $b$ is assigned to a cell $c$ the following values are stored in c:
\begin{align}
c_x &= (x_b - x_c) / w_c & c_y &= (y_b - y_c) / h_c\\
c_w &= w_b / w_c & c_h &= h_b / h_c
\end{align}
where $x_b$, $y_b$ and $x_c$ $y_c$ correspond to the center coordinates of $b$ and $c$ and $w$ and $h$ denote width and hight. Note, that $w_c$ and $h_c$ are always $32$, as the cells of our model have a fixed width and height. We use L1 as our loss
\begin{multline}
\mathrm{loss_{cell}}(c, \hat c) := \delta{c_p} \cdot (|c_x - \hat c_x| + |c_y - \hat c_y| + \\ |c_w - \hat c_w| + |c_w - \hat c_w|)
\label{eq:box}
\end{multline}
where $\hat c$ is the prediction of a cell and $c$ its ground-truth, and $c_p$ denotes whether a positive label has been assigned to a cell. The $\delta{c_p}$ term ensures that the regression loss is zero if no object is present.
We train the confidence labels  using cross-entropy loss. The loss per cell is given as the weighted sum over the confidence and the regression loss. The loss per image is  the mean over the losses of all cells. The KITTI Dataset also contains 'don't Care areas'. Those areas are handled by multiplying the loss of the corresponding cells with zero.
We note, that our label representation is much simpler than Faster-RCNN or ReInspect. This is an additional feature of our detection system.
The loss for MultiNet is given as the sum of the losses for segmentation, detection and classification.

\begin{figure*}[t]
    \begin{subfigure}[t]{0.33\textwidth}
    \centering
        \includegraphics[width=\textwidth]{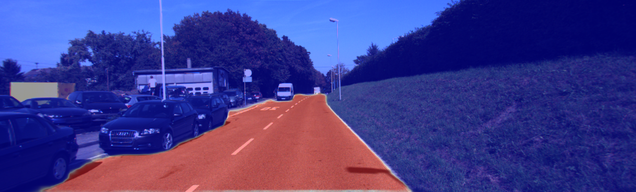}
    \end{subfigure}
    \begin{subfigure}[t]{0.33\textwidth}
    \centering
        \includegraphics[width=\textwidth]{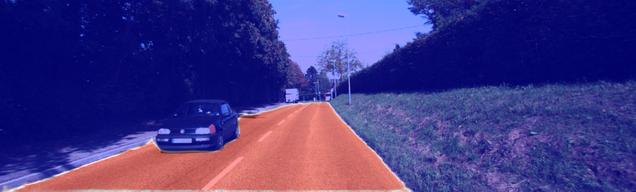}
    \end{subfigure}
    \begin{subfigure}[t]{0.33\textwidth}
        \centering
        \includegraphics[width=\textwidth]{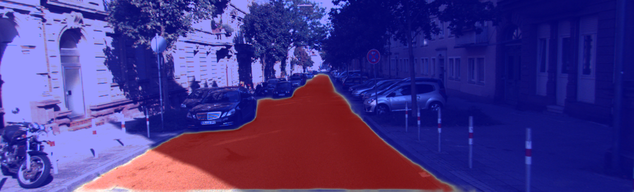}
    \end{subfigure}

    \begin{subfigure}[t]{0.33\textwidth}
    \centering
        \includegraphics[width=\textwidth]{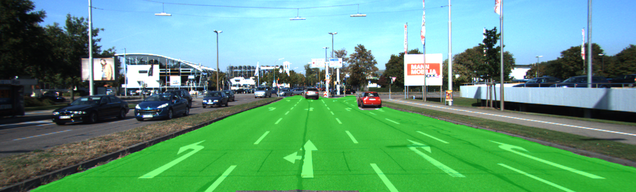}
    \end{subfigure}
    \begin{subfigure}[t]{0.33\textwidth}
    \centering
        \includegraphics[width=\textwidth]{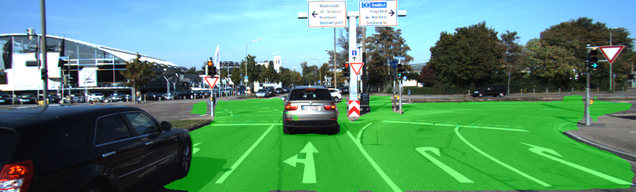}
    \end{subfigure}
    \begin{subfigure}[t]{0.33\textwidth}
    \centering
        \includegraphics[width=\textwidth]{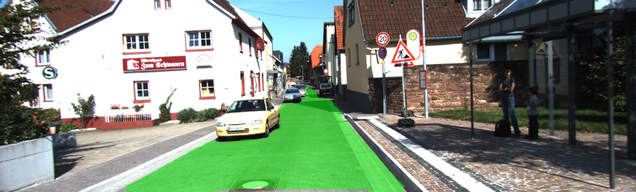}
    \end{subfigure}

    \caption{Visualization of the segmentation output. Top row: Soft segmentation output as red blue plot. The intensity of the plot reflects the confidence. Bottom row hard class labels.}
    \label{fig:seg_output}
\end{figure*} 

The loss for the joint training is given as the sum of the losses for segmentation, detection and classification.

\paragraph{Initialization:} The weights of the encoder are initialized using weights trained  on ImageNet \cite{5206848} data.  
The weights of the detection and classification decoder are initialized using the  initialization scheme of \cite{DBLP:journals/corr/HeZR015}. The  transposed convolution layers of the segmentation decoder are initialized to perform bilinear upsampling. The skip connections of the segmentation decoder are initialized to very small weights. 
Both these modifications  greatly improve segmentation performance.

\paragraph{Optimizer and regularization:}

We use the Adam optimizer \cite{DBLP:journals/corr/KingmaB14} with a learning rate of $10^{-5}$  to train our MultiNet. A weight decay of $5 \cdot 10^{-4}$ is applied to all layers and dropout with probability $0.5$ is applied to the $3 \times 3$ convolution of the classification and all $1 \times 1$ convolutions of the detection decoder.

Standard data augmentation are applied to increase the amount of effective available training data. We augment colour features by applying random brightness and random contrast. Spatial feature are distorted by applying random flip, random resize and random crop.

\section{Experimental Results}
\label{sec:results}

In this section we perform our experimental evaluation on the challenging KITTI dataset. 

\begin{table}[t]
\centering
\begin{tabular}{l | c c | c}
\toprule
Method                     & MaxF1 & AP & Place\\
\midrule
FTP \cite{7535374} & \res{91.61} & \res{90.96} & 6\textsuperscript{th}\\
DDN \cite{1411.4101} & \res{93.43} & \res{89.67} & 5\textsuperscript{th}\\
Up\_Conv\_Poly \cite{Oliveira2016iros} & \res{93.83} & \res{90.47} & 4\textsuperscript{rd}\\
DEEP-DIG \cite{munoz-bulnes_deep_2017} & \res{93.83} & \res{90.47} & 3\textsuperscript{th} \\ 
LoDNN \cite{caltagirone2017fast} & \res{94.07} & \res{92.03} & 2\textsuperscript{rd}\\ %
MultiNet    & \bf{94.88}\% & \bf{93.71}\% & 1\textsuperscript{st} \\ 
\bottomrule
\end{tabular}
\caption{Summary of the URBAN ROAD scores on the public \kitti Road Detection Leaderboard \cite{kitti_bench}.}
\label{tab:kitti_val}
\end{table}

\subsection{Dataset}
We evaluate MultiNet on the KITTI Vision Benchmark Suite \cite{Geiger2013IJRR}. The Benchmark  contains images showing a variety of street situations captured from a moving platform driving around the city of Karlsruhe. 
In addition to the raw data, KITTI  comes with a number of labels for different tasks relevant to autonomous driving. 
We use the road benchmark of \cite{Fritsch2013ITSC}  to evaluate the performance of our semantic segmentation decoder and the object detection benchmark \cite{Geiger2012CVPR} for the detection decoder. 
We exploit the automatically generated labels of  \cite{ma2016find}, which provide us with road labels generated by combining GPS information with open-street map data. 


Detection performance is measured using the average precision score \cite{pascal-voc-2012}. For evaluation, objects are divided into three categories: easy, moderate and hard to detect. 
The segmentation performance is measured using the MaxF1 score \cite{Fritsch2013ITSC}. In addition, the average precision score is given for reference. Classification performance is evaluated by computing the mean accuracy, precision and recall.

\subsection{Experimental evaluation}

The section is structured as fellows. We first evaluate the performance of the three decoders individually. To do this we fine-tune the encoder using just one of the three losses segmentation, detection and classification and compare their performance with a variety of baseline. In the second part we compare joint training of all three decoders with individual inference and show, that the performance of joint training can keep up with the performance of individual inferences. Overall we show, that our approach is competitive with individual inference. This makes our approach very relevant. Joint training has many advantages in robotics application, such as a fast inference time.

\begin{table}[t]
\centering
\begin{tabular}{l | r r }
\toprule
 Task: Metric & MaxF1 & AP \\
\midrule
VGG-pool5 & \res{95.80}  & \res{92.19} \\
ResNet50 & \res{95.89} & \res{92.10}\\
VGG-fc7 & \res{95.94} & \res{92.24} \\
ResNet101 & \bres{96.29} & \bres{92.32} \\
\bottomrule
\end{tabular}
\caption{Performance of the segmentation decoder.}
\label{tab:kitti_res}
\end{table}

\begin{table}[t]
\centering
\begin{tabular}{l | r r r}
\toprule
 Task: Metric & moderate & easy & hard \\
\midrule
VGG no RIO pool & \res{77.00} & \res{86.45} & \res{60.82} \\
Faster-RCNN & \res{78.42} & \res{91.62} & \res{66.85} \\
VGG-pool5 & \res{84.76} & \res{92.18} & \res{68.23} \\
ResNet50 & \res{86.63} & \res{95.55} & \res{74.61} \\
ResNet101 & \bres{89.79} & \bres{96.13} & \bres{77.65} \\
\bottomrule
\end{tabular}
\caption{Performance of our detection decoder.}
\label{tab:eval_results}
\end{table}

\paragraph{Segmentation:}

The segmentation decoder encoder is trained using the four different encoders discussed in Section \ref{sec:encoder}. The scores, computed on a halt-out validation set is reported in Table \ref{tab:kitti_val}.

To compare my approach against the state-of-the-art we trained a segmentation network with VGG-fc7 encoder on the whole training set and submitted the results to the KITTI road leaderboard. At submission time my approach archived first place in the benchmark. Recently my approach was overtaken by newer submissions. All non-anonymous submissions to the benchmark are shown in Table \ref{tab:kitti_res}.

Qualitative results are shown in Fig. \ref{fig:seg_output} both as red blue plot showing the confidence level at each pixel as well as a hard prediction using a threshold of $0.5$.

\begin{figure*}[t]
    \begin{subfigure}[t]{0.33\textwidth}
    \centering
        \includegraphics[width=\textwidth]{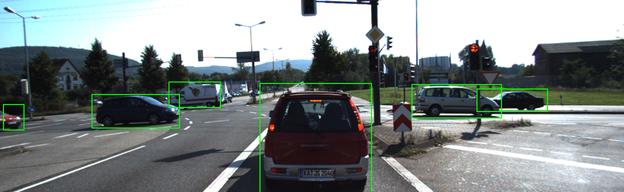}
    \end{subfigure}
    \begin{subfigure}[t]{0.33\textwidth}
    \centering
        \includegraphics[width=\textwidth]{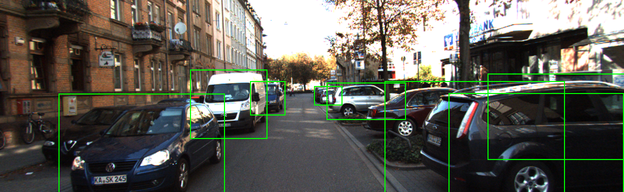}
    \end{subfigure}
    \begin{subfigure}[t]{0.33\textwidth}
        \centering
        \includegraphics[width=\textwidth]{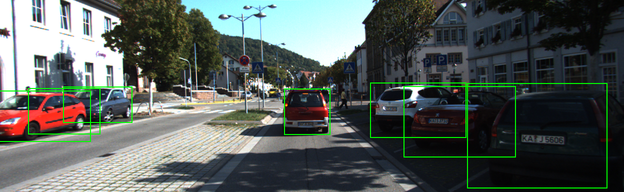}
    \end{subfigure}

    \begin{subfigure}[t]{0.33\textwidth}
        \centering
        \includegraphics[width=\textwidth]{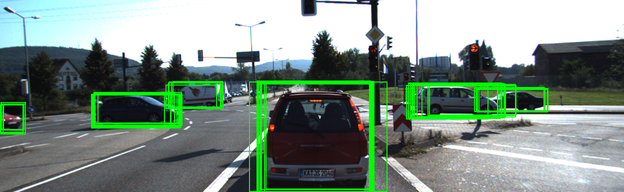}
    \end{subfigure}
    \begin{subfigure}[t]{0.33\textwidth}
    \centering
        \includegraphics[width=\textwidth]{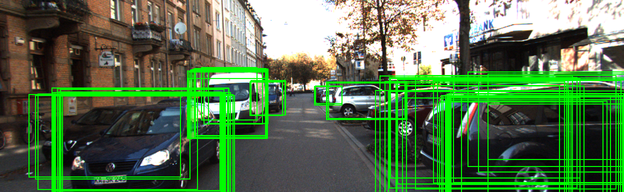}
    \end{subfigure}
    \begin{subfigure}[t]{0.33\textwidth}
    \centering
        \includegraphics[width=\textwidth]{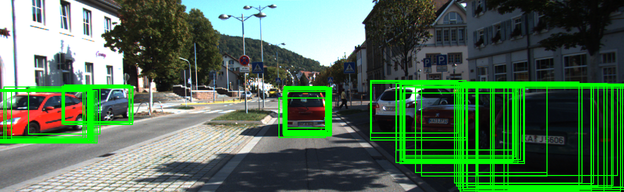}
    \end{subfigure} \vspace{0.1em}

    \begin{subfigure}[t]{0.33\textwidth}
    \centering
        \includegraphics[width=\textwidth]{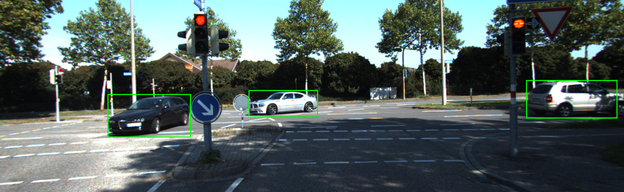}
    \end{subfigure}
    \begin{subfigure}[t]{0.33\textwidth}
    \centering
        \includegraphics[width=\textwidth]{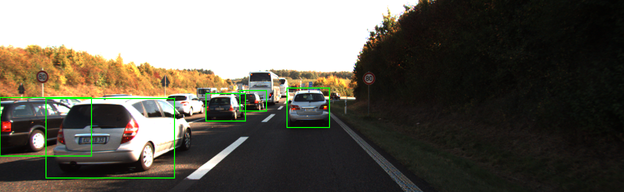}
    \end{subfigure}
    \begin{subfigure}[t]{0.33\textwidth}
    \centering
        \includegraphics[width=\textwidth]{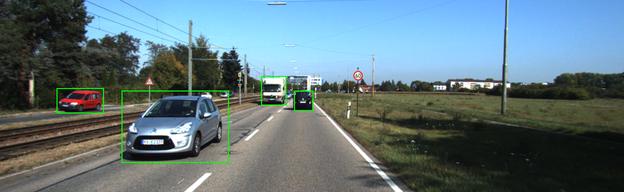}
    \end{subfigure}

    \begin{subfigure}[t]{0.33\textwidth}
        \centering
        \includegraphics[width=\textwidth]{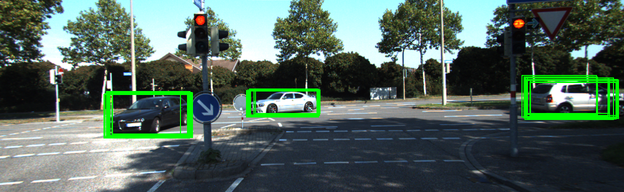}
    \end{subfigure}
    \begin{subfigure}[t]{0.33\textwidth}
    \centering
        \includegraphics[width=\textwidth]{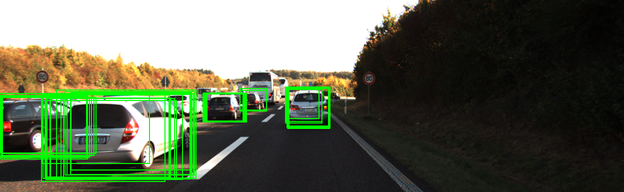}
    \end{subfigure}
    \begin{subfigure}[t]{0.33\textwidth}
        \centering
        \includegraphics[width=\textwidth]{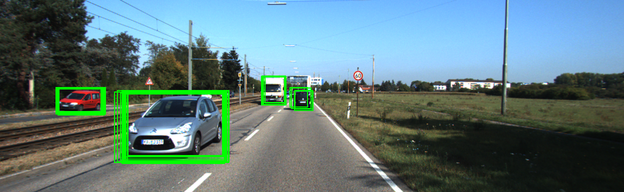}
    \end{subfigure}

    \caption{Visualization of the detection output. With and without non-maximal suppression applied.}
    \label{fig:det_output}
\end{figure*}

\begin{table}[t]
\centering
\begin{tabular}{l| r  r r }
\toprule
            & speed [msec] & speed [fps] \\
\midrule
VGG-pool5 & \bf \SI{42.14}{\ms} & \bf \SI{23.73}{\hertz} \\
ResNet50     & \SI{39.56}{\ms} & \SI{25.27}{\hertz} \\ 
VGG-fc7   & \SI{96.84}{\ms} & \SI{10.32}{\hertz} \\
ResNet101    &  \SI{69.91}{\ms}  & \SI{14.30}{\hertz} \\
\bottomrule
\end{tabular}
\caption{Inference speed of our segmentation.}
\label{tab:seg_speed}
\end{table} 

 \begin{table}[t]
\centering
\begin{tabular}{l| r  r r }
\toprule
            & speed [msec] & speed [fps] & processing \\
\midrule
VGG no RIO & \bf \SI{35.75}{\ms} & \bf \SI{27.96}{\hertz} & \SI{2.46}{\ms} \\
Faster-RCNN & \SI{78.42}{\ms} & \SI{12.75}{\hertz} & \SI{5.3}{\ms} \\
VGG-pool5    & \SI{37.31}{\ms} & \SI{26.79}{\hertz}  & \SI{3.61}{\ms} \\
ResNet50    & \SI{40.09}{\ms} & \SI{24.93}{\hertz} & \SI{3.19}{\ms} \\
ResNet101    & \SI{65.89}{\ms}  & \SI{15.17}{\ms} & \SI{3.11}{\ms} \\
\bottomrule
\end{tabular}
\caption{Inference speed of our detection decoder.}
\label{tab:fast_speed}
\end{table}

\paragraph{Detection:} 

The detection decoder is trained and evaluated on the data provided by the KITTI object benchmark \cite{Geiger2012CVPR}. We train the detection decoder on a VGG \cite{Simonyan14c} and ResNet \cite{DBLP:journals/corr/HeZRS15} decoder and evaluate on a validation set. Table \ref{tab:eval_results} shows the results of our decoder compared to a Faster-RCNN baseline, evaluated on the same validation set. The results show that our rescaling approach is very efficient. Training the detection decoder with rescaling is only marginality slower then training it without. However it offers a significant improvement in detection performance. Overall our approach archives is speed-up over faster-rcnn of almost a factor 2 and outperforms its detection accuracy. Qualitative results of the detection decoder can be seen in \ref{fig:det_output}. 

All in all my results indicate that utilizing a rescaling layer in order to archive scale invariance is a good idea. A rescaling layer might be the key to closing the gap between proposal and non-proposal based approaches.

Our detection decoder is trained and evaluated on the data provided by the KITTI object benchmark \cite{Geiger2012CVPR}. We train our detection decoder on a VGG \cite{Simonyan14c} and ResNet \cite{DBLP:journals/corr/HeZRS15} decoder and evaluate on a validation set. Table \ref{tab:eval_results} shows the results of our decoder compared to a Faster-RCNN baseline, evaluated on the same validation set. We report the inference speed in Table \ref{tab:fast_speed}. We observe that our approach archives is speed-up over faster-rcnn of almost a factor 2 and outperforms its detection accuracy. This makes our decoder particularly suitable for real-time applications. Qualitative results of our detection decoder can be seen in \ref{fig:det_output}.

 \begin{table}[t]
\centering
\begin{tabular}{l | r r r}
\toprule
 & mean Acc. & Precision & Recall \\
\midrule
VGG pool5 [our] & \res{97.34}   & \res{98.52} & \res{87.58} \\
ResNet50  [our] & \res{98.86}   & \bres{100.00} & \res{94.11} \\
ResNet101 [our] & \bres{99.84}   & \res{98.70} & \bres{100.00} \\
\midrule
VGG16 [base] & \res{93.04}       & \res{91.61} & \res{87.90} \\
ResNet101 [base]& \res{93.83}   & \res{91.94} & \res{89.54} \\
\bottomrule
\end{tabular}
\caption{Classification performance of our decoder compared to baseline classification.}
\label{tab:class_results}
\end{table}

\begin{figure*}
    \begin{subfigure}[t]{0.33\textwidth}
        \centering
        \includegraphics[width=\textwidth]{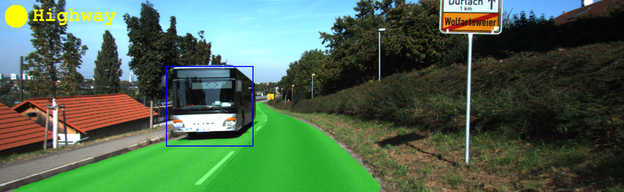}
    \end{subfigure}
    \begin{subfigure}[t]{0.33\textwidth}
    \centering
        \includegraphics[width=\textwidth]{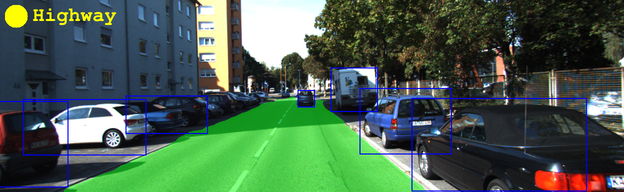}
    \end{subfigure}
    \begin{subfigure}[t]{0.33\textwidth}
    \centering
    \includegraphics[width=\textwidth]{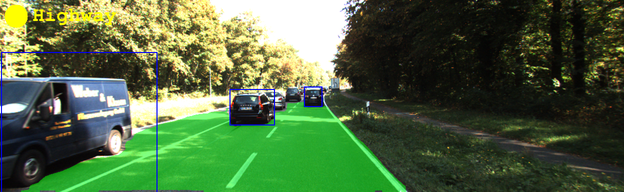}
    
    \end{subfigure}

    \begin{subfigure}[t]{0.33\textwidth}
        \centering
        \includegraphics[width=\textwidth]{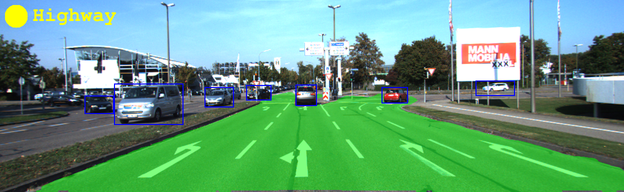}
    \end{subfigure}
    \begin{subfigure}[t]{0.33\textwidth}
    \centering
        \includegraphics[width=\textwidth]{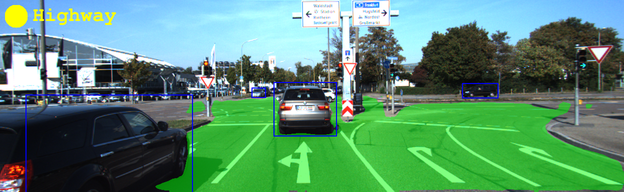}
    \end{subfigure}
    \begin{subfigure}[t]{0.33\textwidth}
    \centering
        \includegraphics[width=\textwidth]{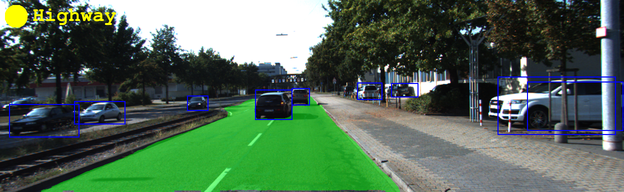}
    \end{subfigure}

    \begin{subfigure}[t]{0.33\textwidth}
    \centering
    \includegraphics[width=\textwidth]{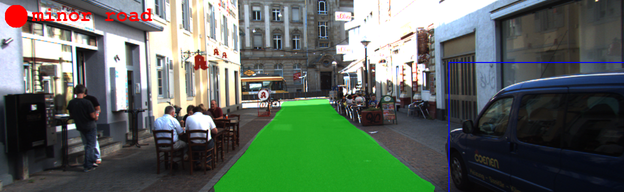}
    \end{subfigure}
    \begin{subfigure}[t]{0.33\textwidth}
    \centering
        \includegraphics[width=\textwidth]{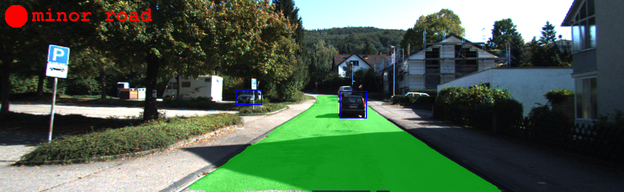}
    \end{subfigure}
    \begin{subfigure}[t]{0.33\textwidth}
    \centering
        \includegraphics[width=\textwidth]{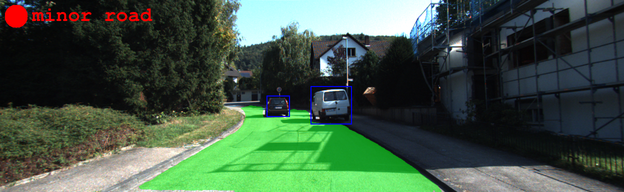}
    \end{subfigure}

    \begin{subfigure}[t]{0.33\textwidth}
    \centering
        \includegraphics[width=\textwidth]{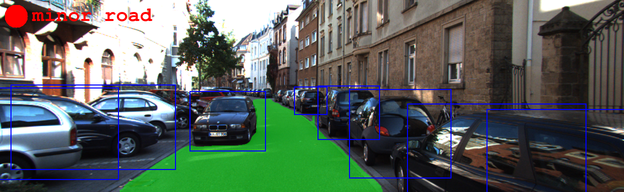}
    \end{subfigure}
    \begin{subfigure}[t]{0.33\textwidth}
    \centering
        \includegraphics[width=\textwidth]{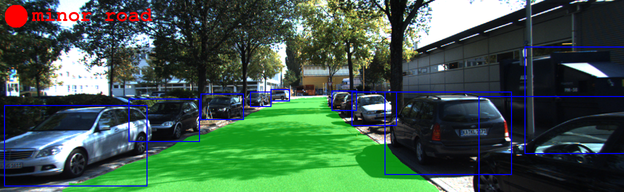}
    \end{subfigure}
    \begin{subfigure}[t]{0.33\textwidth}
    \centering
        \includegraphics[width=\textwidth]{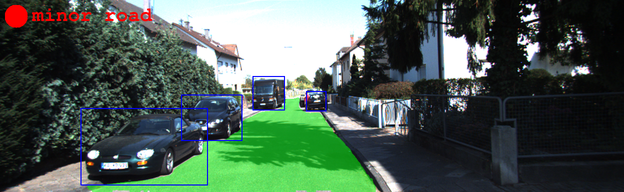}
    \end{subfigure}

    \begin{subfigure}[t]{0.33\textwidth}
    \centering
        \includegraphics[width=\textwidth]{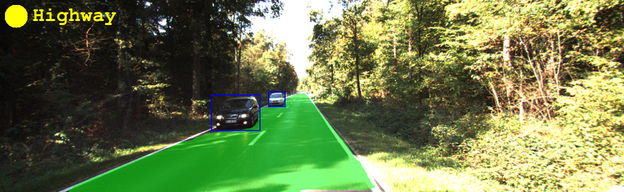}
    \end{subfigure}
    \begin{subfigure}[t]{0.33\textwidth}
    \centering
        \includegraphics[width=\textwidth]{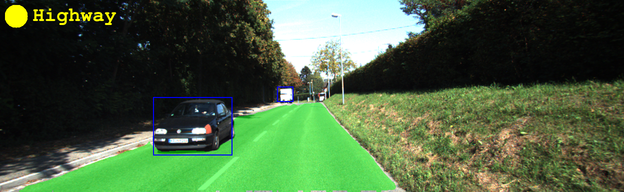}
    \end{subfigure}
    \begin{subfigure}[t]{0.33\textwidth}
        \centering
        \includegraphics[width=\textwidth]{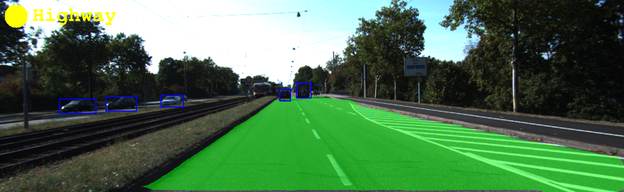}
    \end{subfigure}
    \begin{subfigure}[t]{0.33\textwidth}
        \centering
        \includegraphics[width=\textwidth]{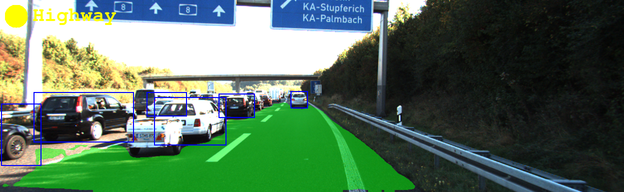}
    \end{subfigure}
    \begin{subfigure}[t]{0.33\textwidth}
    \centering
        \includegraphics[width=\textwidth]{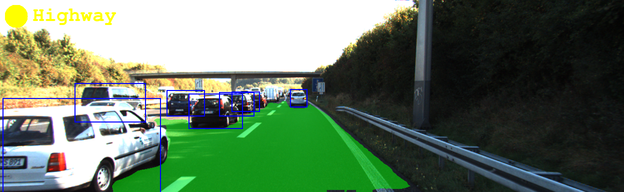}
    \end{subfigure}
    \begin{subfigure}[t]{0.33\textwidth}
    \centering
        \includegraphics[width=\textwidth]{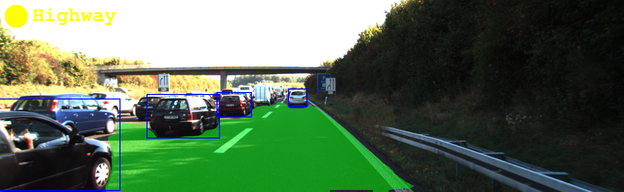}
    \end{subfigure}

    \caption{Visualization of the MultiNet output.}
    \label{fig:multinet_output}
\end{figure*}

\paragraph{Classification:}

The classification data is not part of the official KITTI Benchmark. To evaluate the classification decoder we first need to create our own dataset. This is done using the method descriped in \cite{ma2016find}. To obtain a meaningful task all images of one scene ether fully in the train or fully in the validation set. This is important as the images of one scene are usually visually very similar.

We use a vanilla ResNet and VGG classification approach as baseline and compare this to a VGG and ResNet approach with my classification decoder. The differences between those two approaches are discussed in more detail in Section \ref{sec:class_decoder}. The results are reported in Table \ref{tab:class_results} and Table \ref{tab:class_speed}. Our customised classification decoder clearly outperforms vanilla decoders, showing the effectiveness of my approach.

\begin{table}[t]
\centering
\begin{tabular}{l| r  r r }
\toprule
            & speed [msec] & speed [fps] \\
\midrule
VGG pool5 [our] & \bf \SI{37.83}{\ms} & \bf \SI{26.43}{\hertz} \\
ResNet50  [our] & \SI{44.27}{\ms} & \SI{27.96}{\hertz} \\
ResNet101 [our] & \SI{71.62}{\ms} & \SI{22.58}{\hertz} \\ 
\midrule
VGG16 [base]    & \SI{7.10}{\ms} & \SI{140}{\hertz} \\
ResNet101 [base]& \SI{33.06}{\ms} & \SI{30.24}{\hertz} \\
\bottomrule
\end{tabular}
\caption{Inference speed of our classification.}
\label{tab:class_speed}
\end{table}

\begin{table*}[t]
\centering
\begin{tabular}{l | r r | r r r | r r r}
\toprule
         & MaxF1 & AP & moderate & easy & hard & m. Acc. & Precision & Recall \\
\midrule
VGG pool5 & \res{95.99} & \res{92.31} & \res{84.68} & \res{92.06} & \res{72.08} & \res{95.75} & \res{100} & \res{91.50} \\
ResNet50 & \bres{96.35} & \res{92.13} & \res{86.92} & \bres{96.84} & \res{72.75}  & \res{98.36} & \res{100} & \res{96.73} \\
ResNet101 & \res{95.99} & \bres{91.99} & \bres{89.30} & \res{96.31} & \bres{75.42}  & \bres{98.61} & \res{99.33} & \bres{97.38} \\
\bottomrule
\end{tabular}
\caption{Results of joint training}
\label{tab:multinet_results}
\end{table*}

\begin{table}[t]
\centering
\begin{tabular}{l| r  r r }
\toprule
            & speed [msec] & speed [fps] \\
\midrule
VGG pool5 & \bf \SI{42.48}{\ms} & \bf \SI{23.53}{\hertz}\\
ResNet50 & \SI{60.22}{\ms} & \SI{16.60}{\hertz}\\
ResNet101 & \SI{79.70}{\ms} & \SI{12.54}{\hertz}\\
\bottomrule
\end{tabular}
\caption{Speed of joint inference.}
\label{tab:multinet_speed}
\end{table}

\paragraph{MultiNet:}

We ran a series of experiments comparing VGG and ResNet as encoder. Table \ref{tab:multinet_results} and Table \ref{tab:multinet_speed} compare performance of VGG and ResNet. We observe, that both ResNet-based encoders are able to outperform VGG slightly. There is however a trade-off, as  the VGG encoder is faster.

The speed gap between VGG pool5 and ResNet50 is much larger when performing joint inference compared to the individual task. This can be explained by the fact that ResNet computes features with $2048$ channels, while VGG features have only $512$ channels. Thus, computing the fist layer of each decoder is significantly more expensive.

Overall we conclude, that MultiNet using a VGG decoder offers a very good trade-off between performance and speed.

\section{Conclusion}
\label{sec:conclusion}

In this paper we have developed a unified deep architecture which is able to jointly reason about 
classification, detection and semantic segmentation.  
Our approach is very simple, can be trained end-to-end and performs extremely well in the challenging KITTI dataset, outperforming the state-of-the-art in  the road segmentation task.
Our approach is also very efficient, taking \SI{42.48}{\ms} to perform all tasks. In the future we plan to exploit compression methods in order to further reduce the computational bottleneck and energy consumption of MutiNet.

%
%


\paragraph{Acknowledgements: } This work was partially supported by {Begabtenstiftung Informatik Karlsruhe}, 
ONR-N00014-14-1-0232, Qualcomm, Samsung, NVIDIA, Google, EPSRC and NSERC. We are thankful to Thomas Roddick for proofreading the paper.

{\small
\bibliographystyle{ieee}
\bibliography{multinet}

\begin{thebibliography}{10}\itemsep=-1pt

\bibitem{DBLP:journals/corr/BadrinarayananK15}
V.~Badrinarayanan, A.~Kendall, and R.~Cipolla.
\newblock Segnet: {A} deep convolutional encoder-decoder architecture for image
  segmentation.
\newblock {\em CoRR}, abs/1511.00561, 2015.

\bibitem{caltagirone2017fast}
L.~Caltagirone, S.~Scheidegger, L.~Svensson, and M.~Wahde.
\newblock Fast lidar-based road detection using convolutional neural networks.
\newblock {\em arXiv preprint arXiv:1703.03613}, 2017.

\bibitem{CRF2}
L.~Chen, G.~Papandreou, I.~Kokkinos, K.~Murphy, and A.~L. Yuille.
\newblock Semantic image segmentation with deep convolutional nets and fully
  connected crfs.
\newblock {\em CoRR}, abs/1412.7062, 2014.

\bibitem{DBLP:journals/corr/ChenPK0Y16}
L.~Chen, G.~Papandreou, I.~Kokkinos, K.~Murphy, and A.~L. Yuille.
\newblock Deeplab: Semantic image segmentation with deep convolutional nets,
  atrous convolution, and fully connected crfs.
\newblock {\em CoRR}, abs/1606.00915, 2016.

\bibitem{chen2016monocular}
X.~Chen, K.~Kundu, Z.~Zhang, H.~Ma, S.~Fidler, and R.~Urtasun.
\newblock Monocular 3d object detection for autonomous driving.
\newblock In {\em Proceedings of the IEEE Conference on Computer Vision and
  Pattern Recognition}, pages 2147--2156, 2016.

\bibitem{chen20153d}
X.~Chen, K.~Kundu, Y.~Zhu, A.~G. Berneshawi, H.~Ma, S.~Fidler, and R.~Urtasun.
\newblock 3d object proposals for accurate object class detection.
\newblock In {\em Advances in Neural Information Processing Systems}, pages
  424--432, 2015.

\bibitem{dai2016instance}
J.~Dai, K.~He, and J.~Sun.
\newblock Instance-aware semantic segmentation via multi-task network cascades.
\newblock In {\em Proceedings of the IEEE Conference on Computer Vision and
  Pattern Recognition}, pages 3150--3158, 2016.

\bibitem{5206848}
J.~Deng, W.~Dong, R.~Socher, L.-J. Li, K.~Li, and L.~Fei-Fei.
\newblock Imagenet: A large-scale hierarchical image database.
\newblock In {\em Computer Vision and Pattern Recognition, 2009. CVPR 2009.
  IEEE Conference on}, pages 248--255, June 2009.

\bibitem{dumoulin2016guide}
V.~Dumoulin and F.~Visin.
\newblock A guide to convolution arithmetic for deep learning.
\newblock {\em arXiv preprint arXiv:1603.07285}, 2016.

\bibitem{DBLP:journals/corr/ErhanSTA13}
D.~Erhan, C.~Szegedy, A.~Toshev, and D.~Anguelov.
\newblock Scalable object detection using deep neural networks.
\newblock {\em CoRR}, abs/1312.2249, 2013.

\bibitem{pascal-voc-2012}
M.~Everingham, L.~Van~Gool, C.~K.~I. Williams, J.~Winn, and A.~Zisserman.
\newblock The {PASCAL} {V}isual {O}bject {C}lasses {C}hallenge 2012 {(VOC2012)}
  {R}esults.
\newblock
  http://www.pascal-network.org/challenges/VOC/voc2012/workshop/index.html.

\bibitem{Fritsch2013ITSC}
J.~Fritsch, T.~Kuehnl, and A.~Geiger.
\newblock A new performance measure and evaluation benchmark for road detection
  algorithms.
\newblock In {\em International Conference on Intelligent Transportation
  Systems (ITSC)}, 2013.

\bibitem{kitti_bench}
A.~Geiger.
\newblock Kitti road public benchmark, 2013.

\bibitem{Geiger2013IJRR}
A.~Geiger, P.~Lenz, C.~Stiller, and R.~Urtasun.
\newblock Vision meets robotics: The kitti dataset.
\newblock {\em International Journal of Robotics Research (IJRR)}, 2013.

\bibitem{Geiger2012CVPR}
A.~Geiger, P.~Lenz, and R.~Urtasun.
\newblock Are we ready for autonomous driving? the kitti vision benchmark
  suite.
\newblock In {\em Conference on Computer Vision and Pattern Recognition
  (CVPR)}, 2012.

\bibitem{gidaris2015object}
S.~Gidaris and N.~Komodakis.
\newblock Object detection via a multi-region and semantic segmentation-aware
  cnn model.
\newblock In {\em Proceedings of the IEEE International Conference on Computer
  Vision}, pages 1134--1142, 2015.

\bibitem{DBLP:journals/corr/Girshick15}
R.~B. Girshick.
\newblock Fast {R-CNN}.
\newblock {\em CoRR}, abs/1504.08083, 2015.

\bibitem{DBLP:journals/corr/GirshickDDM13}
R.~B. Girshick, J.~Donahue, T.~Darrell, and J.~Malik.
\newblock Rich feature hierarchies for accurate object detection and semantic
  segmentation.
\newblock {\em CoRR}, abs/1311.2524, 2013.

\bibitem{fast_scanning}
A.~Giusti, D.~C. Ciresan, J.~Masci, L.~M. Gambardella, and J.~Schmidhuber.
\newblock Fast image scanning with deep max-pooling convolutional neural
  networks.
\newblock {\em CoRR}, abs/1302.1700, 2013.

\bibitem{hariharan2014simultaneous}
B.~Hariharan, P.~Arbel{\'a}ez, R.~Girshick, and J.~Malik.
\newblock Simultaneous detection and segmentation.
\newblock In {\em European Conference on Computer Vision}, pages 297--312.
  Springer, 2014.

\bibitem{DBLP:journals/corr/HeGDG17}
K.~He, G.~Gkioxari, P.~Doll{\'{a}}r, and R.~B. Girshick.
\newblock Mask {R-CNN}.
\newblock {\em CoRR}, abs/1703.06870, 2017.

\bibitem{DBLP:journals/corr/HeZRS15}
K.~He, X.~Zhang, S.~Ren, and J.~Sun.
\newblock Deep residual learning for image recognition.
\newblock {\em CoRR}, abs/1512.03385, 2015.

\bibitem{DBLP:journals/corr/HeZR015}
K.~He, X.~Zhang, S.~Ren, and J.~Sun.
\newblock Delving deep into rectifiers: Surpassing human-level performance on
  imagenet classification.
\newblock {\em CoRR}, abs/1502.01852, 2015.

\bibitem{DBLP:journals/corr/HosangBDS15}
J.~H. Hosang, R.~Benenson, P.~Doll{\'{a}}r, and B.~Schiele.
\newblock What makes for effective detection proposals?
\newblock {\em CoRR}, abs/1502.05082, 2015.

\bibitem{DBLP:journals/corr/HosangBS14}
J.~H. Hosang, R.~Benenson, and B.~Schiele.
\newblock How good are detection proposals, really?
\newblock {\em CoRR}, abs/1406.6962, 2014.

\bibitem{DBLP:journals/corr/ImKJM16}
D.~J. Im, C.~D. Kim, H.~Jiang, and R.~Memisevic.
\newblock Generating images with recurrent adversarial networks.
\newblock {\em CoRR}, abs/1602.05110, 2016.

\bibitem{DBLP:journals/corr/KingmaB14}
D.~P. Kingma and J.~Ba.
\newblock Adam: {A} method for stochastic optimization.
\newblock {\em CoRR}, abs/1412.6980, 2014.

\bibitem{cifar}
A.~Krizhevsky, V.~Nair, and G.~Hinton.
\newblock Cifar-10 (canadian institute for advanced research).

\bibitem{NIPS2012_4824}
A.~Krizhevsky, I.~Sutskever, and G.~E. Hinton.
\newblock Imagenet classification with deep convolutional neural networks.
\newblock In F.~Pereira, C.~J.~C. Burges, L.~Bottou, and K.~Q. Weinberger,
  editors, {\em Advances in Neural Information Processing Systems 25}, pages
  1097--1105. Curran Associates, Inc., 2012.

\bibitem{7535374}
A.~Laddha, M.~K. Kocamaz, L.~E. Navarro-Serment, and M.~Hebert.
\newblock Map-supervised road detection.
\newblock In {\em 2016 IEEE Intelligent Vehicles Symposium (IV)}, pages
  118--123, June 2016.

\bibitem{lampert2008beyond}
C.~H. Lampert, M.~B. Blaschko, and T.~Hofmann.
\newblock Beyond sliding windows: Object localization by efficient subwindow
  search.
\newblock In {\em Computer Vision and Pattern Recognition, 2008. CVPR 2008.
  IEEE Conference on}, pages 1--8. IEEE, 2008.

\bibitem{highly}
H.~Li, R.~Zhao, and X.~Wang.
\newblock Highly efficient forward and backward propagation of convolutional
  neural networks for pixelwise classification.
\newblock {\em CoRR}, abs/1412.4526, 2014.

\bibitem{DBLP:journals/corr/LiuAESR15}
W.~Liu, D.~Anguelov, D.~Erhan, C.~Szegedy, and S.~E. Reed.
\newblock {SSD:} single shot multibox detector.
\newblock {\em CoRR}, abs/1512.02325, 2015.

\bibitem{liu2015representation}
X.~Liu, J.~Gao, X.~He, L.~Deng, K.~Duh, and Y.-Y. Wang.
\newblock Representation learning using multi-task deep neural networks for
  semantic classification and information retrieval.
\newblock In {\em Proc. NAACL}, 2015.

\bibitem{long2015fully}
J.~Long, E.~Shelhamer, and T.~Darrell.
\newblock Fully convolutional networks for semantic segmentation.
\newblock In {\em Proceedings of the IEEE Conference on Computer Vision and
  Pattern Recognition}, pages 3431--3440, 2015.

\bibitem{DBLP:journals/corr/Long015a}
M.~Long and J.~Wang.
\newblock Learning multiple tasks with deep relationship networks.
\newblock {\em CoRR}, abs/1506.02117, 2015.

\bibitem{ma2016find}
W.-C. Ma, S.~Wang, M.~A. Brubaker, S.~Fidler, and R.~Urtasun.
\newblock Find your way by observing the sun and other semantic cues.
\newblock {\em arXiv preprint arXiv:1606.07415}, 2016.

\bibitem{1411.4101}
R.~Mohan.
\newblock Deep deconvolutional networks for scene parsing, 2014.

\bibitem{munoz-bulnes_deep_2017}
J.~Muñoz-Bulnes, C.~Fernandez, I.~Parra, D.~Fernández-Llorca, and M.~A.
  Sotelo.
\newblock Deep {Fully} {Convolutional} {Networks} with {Random} {Data}
  {Augmentation} for {Enhanced} {Generalization} in {Road} {Detection}.
\newblock In {\em Submitted to the Workshop on {Deep} {Learning} for
  {Autonomous} {Driving} on {IEEE} 20th {International} {Conference} on
  {Intelligent} {Transportation} {Systems}}, Yokohama, Japan, Oct. 2017.

\bibitem{noh2015learning}
H.~Noh, S.~Hong, and B.~Han.
\newblock Learning deconvolution network for semantic segmentation.
\newblock 2015.

\bibitem{Oliveira2016iros}
G.~Oliveira, W.~Burgard, and T.~Brox.
\newblock Efficient deep methods for monocular road segmentation.
\newblock 2016.

\bibitem{googleSeg}
G.~Papandreou, L.~Chen, K.~Murphy, and A.~L. Yuille.
\newblock Weakly- and semi-supervised learning of a {DCNN} for semantic image
  segmentation.
\newblock {\em CoRR}, abs/1502.02734, 2015.

\bibitem{pinheiro2016learning}
P.~O. Pinheiro, T.-Y. Lin, R.~Collobert, and P.~Doll{\'a}r.
\newblock Learning to refine object segments.
\newblock In {\em European Conference on Computer Vision}, pages 75--91.
  Springer, 2016.

\bibitem{DBLP:journals/corr/RanjanPC16}
R.~Ranjan, V.~M. Patel, and R.~Chellappa.
\newblock Hyperface: {A} deep multi-task learning framework for face detection,
  landmark localization, pose estimation, and gender recognition.
\newblock {\em CoRR}, abs/1603.01249, 2016.

\bibitem{DBLP:journals/corr/RedmonDGF15}
J.~Redmon, S.~K. Divvala, R.~B. Girshick, and A.~Farhadi.
\newblock You only look once: Unified, real-time object detection.
\newblock {\em CoRR}, abs/1506.02640, 2015.

\bibitem{DBLP:journals/corr/RenHG015}
S.~Ren, K.~He, R.~B. Girshick, and J.~Sun.
\newblock Faster {R-CNN:} towards real-time object detection with region
  proposal networks.
\newblock {\em CoRR}, abs/1506.01497, 2015.

\bibitem{DBLP:journals/corr/RonnebergerFB15}
O.~Ronneberger, P.~Fischer, and T.~Brox.
\newblock U-net: Convolutional networks for biomedical image segmentation.
\newblock {\em CoRR}, abs/1505.04597, 2015.

\bibitem{ILSVRC15}
O.~Russakovsky, J.~Deng, H.~Su, J.~Krause, S.~Satheesh, S.~Ma, Z.~Huang,
  A.~Karpathy, A.~Khosla, M.~Bernstein, A.~C. Berg, and L.~Fei-Fei.
\newblock {ImageNet Large Scale Visual Recognition Challenge}.
\newblock {\em International Journal of Computer Vision (IJCV)},
  115(3):211--252, 2015.

\bibitem{DBLP:journals/corr/SchwingU15}
A.~G. Schwing and R.~Urtasun.
\newblock Fully connected deep structured networks.
\newblock {\em CoRR}, abs/1503.02351, 2015.

\bibitem{seegertowards}
C.~Seeger, A.~M{\"u}ller, L.~Schwarz, and M.~Manz.
\newblock Towards road type classification with occupancy grids.
\newblock {\em IVS Workshop}, 2016.

\bibitem{DBLP:journals/corr/SermanetEZMFL13}
P.~Sermanet, D.~Eigen, X.~Zhang, M.~Mathieu, R.~Fergus, and Y.~LeCun.
\newblock Overfeat: Integrated recognition, localization and detection using
  convolutional networks.
\newblock {\em CoRR}, abs/1312.6229, 2013.

\bibitem{Simonyan14c}
K.~Simonyan and A.~Zisserman.
\newblock Very deep convolutional networks for large-scale image recognition.
\newblock {\em CoRR}, abs/1409.1556, 2014.

\bibitem{stewart2016end}
R.~Stewart, M.~Andriluka, and A.~Y. Ng.
\newblock End-to-end people detection in crowded scenes.
\newblock In {\em Proceedings of the IEEE Conference on Computer Vision and
  Pattern Recognition}, pages 2325--2333, 2016.

\bibitem{DBLP:journals/corr/WuSH16e}
Z.~Wu, C.~Shen, and A.~van~den Hengel.
\newblock Wider or deeper: Revisiting the resnet model for visual recognition.
\newblock {\em CoRR}, abs/1611.10080, 2016.

\bibitem{yim2015rotating}
J.~Yim, H.~Jung, B.~Yoo, C.~Choi, D.~Park, and J.~Kim.
\newblock Rotating your face using multi-task deep neural network.
\newblock In {\em Proceedings of the IEEE Conference on Computer Vision and
  Pattern Recognition}, pages 676--684, 2015.

\bibitem{DBLP:journals/corr/YuK15}
F.~Yu and V.~Koltun.
\newblock Multi-scale context aggregation by dilated convolutions.
\newblock {\em CoRR}, abs/1511.07122, 2015.

\bibitem{zeiler2014visualizing}
M.~D. Zeiler and R.~Fergus.
\newblock Visualizing and understanding convolutional networks.
\newblock In {\em European Conference on Computer Vision}, pages 818--833.
  Springer, 2014.

\bibitem{zeiler2010deconvolutional}
M.~D. Zeiler, D.~Krishnan, G.~W. Taylor, and R.~Fergus.
\newblock Deconvolutional networks.
\newblock In {\em Computer Vision and Pattern Recognition (CVPR), 2010 IEEE
  Conference on}, pages 2528--2535. IEEE, 2010.

\bibitem{zhang2014facial}
Z.~Zhang, P.~Luo, C.~C. Loy, and X.~Tang.
\newblock Facial landmark detection by deep multi-task learning.
\newblock In {\em European Conference on Computer Vision}, pages 94--108.
  Springer, 2014.

\bibitem{CRF1}
S.~Zheng, S.~Jayasumana, B.~Romera{-}Paredes, V.~Vineet, Z.~Su, D.~Du,
  C.~Huang, and P.~H.~S. Torr.
\newblock Conditional random fields as recurrent neural networks.
\newblock {\em CoRR}, abs/1502.03240, 2015.

\end{thebibliography}
}

\end{document}